\pdfoutput=1

\documentclass[11pt]{article}

\usepackage[]{acl}

\usepackage{times}
\usepackage{latexsym}

\usepackage[T1]{fontenc}

\usepackage[utf8]{inputenc}

\usepackage{microtype}

\usepackage{xcolor}
\usepackage{bm, amsfonts, amsmath}
\usepackage{algorithm, algpseudocode}
\usepackage{newfloat,multirow,booktabs}
\usepackage{subfig,amssymb}
\usepackage{graphicx}

\newcommand\ie{\textit{i.e.}}
\newcommand\eg{\textit{e.g.}}
\newcommand\etc{\textit{etc}}
\newcommand\wrt{\textit{w.r.t}}

\title{Decomposed Meta-Learning for Few-Shot Named Entity Recognition}

\author{Tingting Ma$^{1}$\footnotemark[2] \footnotemark[1] , 
Huiqiang Jiang$^{2}$\footnotemark[1] , 
Qianhui Wu$^{2}$\footnotemark[1] , 
Tiejun Zhao$^{1}$,
Chin-Yew Lin$^{2}$ \\ 
$^{1}$Harbin Institute of Technology, Harbin, China \\
$^{2}$Microsoft Research Asia \\
\tt hittingtingma@gmail.com \\
\tt \{hjiang, qianhuiwu, cyl\}@microsoft.com, tjzhao@hit.edu.cn 
}

\begin{document}
\maketitle

\renewcommand{\thefootnote}{\fnsymbol{footnote}}
\footnotetext[1]{Equal contributions.}
\footnotetext[2]{Work during internship at Microsoft Research Asia.}

\renewcommand{\thefootnote}{\arabic{footnote}}

\begin{abstract}
Few-shot named entity recognition (NER) systems aim at recognizing novel-class named entities based on only a few labeled examples. In this paper, we present a decomposed meta-learning approach which addresses the problem of few-shot NER by sequentially tackling few-shot span detection and few-shot entity typing using meta-learning. In particular, we take the few-shot span detection as a sequence labeling problem and train the span detector by introducing the model-agnostic meta-learning (MAML) algorithm to find a good model parameter initialization that could fast adapt to new entity classes. For few-shot entity typing, we propose MAML-ProtoNet, \ie, MAML-enhanced prototypical networks to find a good embedding space that can better distinguish text span representations from different entity classes. Extensive experiments on various benchmarks show that our approach achieves superior performance over prior methods.\footnote{Our implementation is publicly available at \url{https://github.com/microsoft/vert-papers/tree/master/papers/DecomposedMetaNER}}
\end{abstract}

\section{Introduction}
Named entity recognition (NER) aims at locating and classifying text spans into pre-defined entity classes such as locations, organizations, \etc.
Deep neural architectures have shown great success in fully supervised NER \citep{lample2016neural,ma2016end,chiu2016named,peters2017semi} with a fair amount of labeled data available for training.
However, in practical applications, NER systems are usually expected to rapidly adapt to some new entity types unseen during training.
It is costly while not flexible to collect a number of additional labeled data for these types.
As a result, the problem of few-shot NER, which involves learning unseen entity types from only a few labeled examples for each class (also known as \emph{support examples}), has attracted considerable attention from the research community in recent years.

Previous studies on few-shot NER are typically based on token-level metric learning, in which a model compares each query token to the prototype~\citep{snell2017proto} of each entity class or each token of support examples and assign the label according to their distances~\citep{fritzler2019few, hou2020few, yang2020simple}. Alternatively, some more recent attempts have switched to span-level metric-learning~\citep{yu2021fewShot,wang2021enhanced} to bypass the issue of token-wise label dependency while explicitly utilizing phrasal representations.


However, these methods based on metric learning might be less effective when encountering large domain gap, since they just directly use the learned metric without any further adaptation to the target domain. In other words, they do not fully explore the information brought by the support examples.
There also exist additional limitations in the current methods based on span-level metric learning.
First, the decoding process 
requires careful handling of
overlapping spans due to the nature of span enumeration.
Second, the class prototype corresponding to non-entities (\ie, prototype of the ``\texttt{O}'' class) is usually noisy because 
non-entity common words in the large vocabulary rarely share anything together in common.
Moreover, when targeting at a different domain, the only available information useful for domain transfer is the limited number of support examples.
Unfortunately, these key examples are only used for inference-phase similarity calculation in previous methods.

To tackle these limitations, this paper presents a decomposed meta-learning framework that addresses the problem of few-shot NER by sequentially conducting \textit{few-shot entity span detection} and \textit{few-shot entity typing} respectively via meta-learning.
Specifically, for \emph{few-shot span detection}, we model it as a sequence labeling problem to avoid handling overlapping spans.
Note that the detection model aims at \emph{locating} named entities and is class-agnostic.
We only feed the detected entity spans to the typing model for entity class inference,
and hence the problem of noisy ``\texttt{O}'' prototype could also be eliminated.
When training the span detector, 
we specifically use the model-agnostic meta-learning (MAML)~\citep{finn2017model} algorithm to find a good model parameter initialization
that could fast adapt to new entity classes
with learned class-agnostic meta-knowledge of span boundaries
after updating with the target-domain support examples.
The boundary information of domain-specific entities from the support examples is supposed to be effectively leveraged via these update steps such that the model could better transfer to the target domain.
For \emph{few-shot entity typing}, we
implement the typing model with standard prototypical networks~\citep[ProtoNet]{snell2017proto},
and propose MAML-ProtoNet to narrow the gap between source domains and the target domain.
Compared with ProtoNet which only uses support examples for inference-phase similarity calculation, the proposed MAML-Proto additionally utilizes these examples to modify the shared embedding space of spans and prototypes by clustering spans representations from the same entity class while dispersing those from different entity classes for more accurate predictions.

We evaluate our proposed framework on several benchmark datasets with different few-shot settings.
Experimental results show that our framework achieves superior performance over previous state-of-the-art methods.
We also conduct qualitative and quantitative analyses 
over how the different strategies to conduct meta-learning might affect the performance.


\begin{figure*}[t]
    \centering
    \includegraphics[width=2\columnwidth]{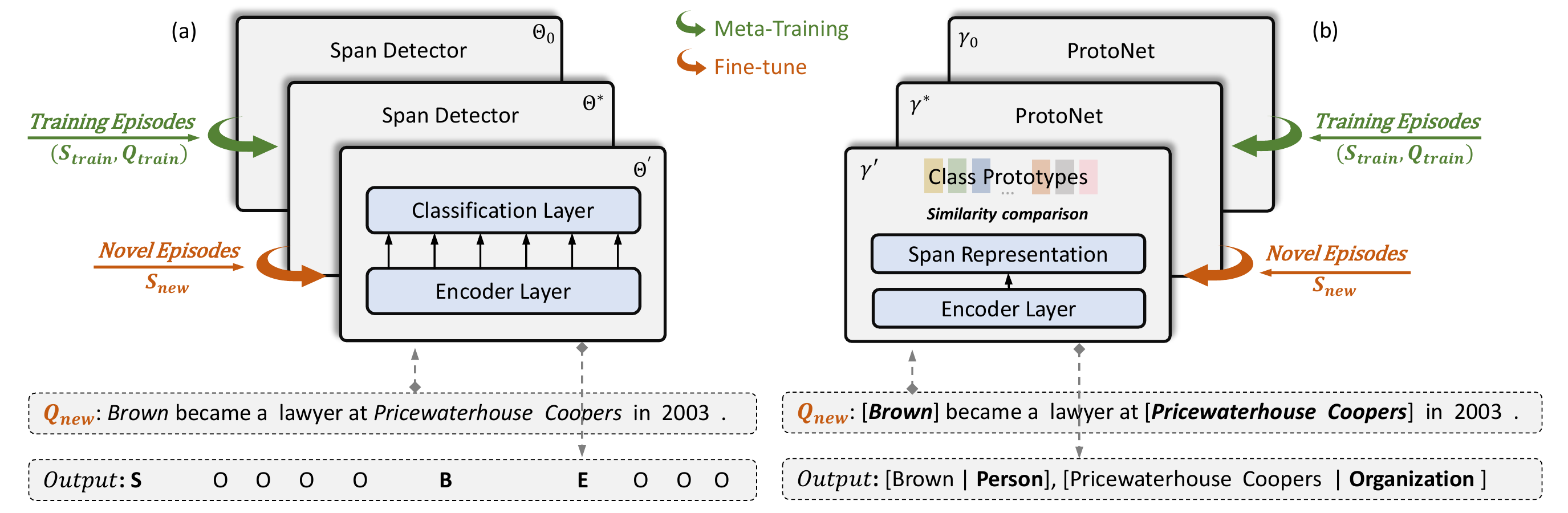}
    \caption{The framework of our proposed approach is decomposed into two modules: (a) entity span detection with parameters $\Theta$ and (b) entity typing with parameters $\gamma$. Two modules are trained independently using ($S_{train}, Q_{train}$). At meta-test time, these two modules firstly are finetuned on the support set $S_{new}$, then given a query sentence in $Q_{new}$, the spans detected by (a) are sent to (b) for entity typing.}
    \label{fig:framework}
\end{figure*}

\section{Task Definition}
Given an input sequence $\bm{x}=\{x_i\}_{i=1}^L$ with $L$ tokens, an NER system is supposed to output a label sequence $\bm{y}=\{y_i\}_{i=1}^L$, where $x_i$ is the $i$-th token, $y_i\in\mathcal{Y}\cup\{\texttt{O}\}$ is the label of $x_i$, $\mathcal{Y}$ is the pre-defined entity class set, and $\texttt{O}$ denotes non-entities.

In this paper, we focus on the standard $N$-way $K$-shot setting
as in \citet{ding2021nerd}. An example of 2-way 1-shot episode is shown in Table \ref{tab:taskexample}.
In the training phase, we consider training episodes $\mathcal{E}_{train}=\{(\mathcal{S}_{train}, \mathcal{Q}_{train}, \mathcal{Y}_{train})\}$ built from source-domain labeled data, where $\mathcal{S}_{train}=\{(\bm{x}^{(i)}, \bm{y}^{(i)})\}_{i=1}^{N\times K}$ denotes the support set, $\mathcal{Q}_{train}=\{\bm{x}^{(j)}, \bm{y}^{(j)}\}_{j=1}^{N\times K'}$ denotes the query set, $\mathcal{Y}_{train}$ denotes the set of entity classes, and $|\mathcal{Y}_{train}|=N$.
In the testing phase, we consider novel episodes $\mathcal{E}_{new}=\{(\mathcal{S}_{new}, \mathcal{Q}_{new}, \mathcal{Y}_{new})\}$ constructed with data from target domains in a similar way.
In the few-shot NER task, a model learned with training episodes $\mathcal{E}_{train}$ is expected to leverage the support set  $\mathcal{S}_{new}=\{(\bm{x}^{(i)}, \bm{y}^{(i)})\}_{i=1}^{N\times K}$ of a novel episode $(\mathcal{S}_{new}, \mathcal{Q}_{new}, \mathcal{Y}_{new}) \in \mathcal{E}_{new}$ to make predictions on the query set $\mathcal{Q}_{new}=\{\bm{x}^{(j)}\}_{j=1}^{N\times K'}$.
Here, $\mathcal{Y}_{new}$ denotes the set of entity classes
with a cardinality of $N$.
Note that, $\forall~\mathcal{Y}_{train}, \mathcal{Y}_{new}$, $\mathcal{Y}_{train}\cap \mathcal{Y}_{new}=\emptyset$.

\begin{table}[htbp]
    \centering
    \small
    \begin{tabular}{l|p{4.45cm}}
    \toprule
    \textbf{Target Types} $\mathcal{Y}$ & \textcolor{blue}{[person-actor]}, \textcolor{red}{[art-film]} \\
    \midrule
    \multirow{5}{*}{\textbf{Support set} $\mathcal{S}$} &
    (1) \emph{\textcolor{blue}{Jack Gordon}}$_{\textcolor{blue}{\textrm{[person-actor]}}}$ ( born 27 June 1985 ) is an English actor .\\
     & (2) This location had also been used to shoot the film `` \emph{\textcolor{red}{Saving Private Ryan}}$_{\textcolor{red}{[\textrm{art-film}]}}$ '' . \\
    \midrule
    \midrule
    \multirow{3}{*}{\textbf{Query Set} $\mathcal{Q}$} &
    Kurland starred in `` Taps '' , which won first prize at the Rhode Island International Film Festival in 2006 . \\
    \midrule
    \multirow{4}{*}{\textbf{Expected output}} & 
    \emph{\textcolor{blue}{Kurland}}$_{\textcolor{blue}{[\textrm{person-actor}]}}$ starred in `` \emph{\textcolor{red}{Taps}}$_{\textcolor{red}{[\textrm{art-film}]}}$ '' , which won first prize at the Rhode Island International Film Festival in 2006 . \\
    \bottomrule
    \end{tabular}
    \caption{An example of the simplest 2-way 1-shot setting, which contains two entity classes and each class has one example (shot) in the support set $\mathcal{S}$. Different colors indicate different entity classes.}
    \label{tab:taskexample}
\end{table}

\section{Methodology}
Figure~\ref{fig:framework} illustrates the overall framework of our decomposed meta-learning approach for few-shot named entity recognition.
It is composed of two steps: \emph{entity span detection} and \emph{entity typing}.

\subsection{Entity Span Detection}
The span detection model aims at locating all the named entities in an input sequence.
The model should be type-agnostic, \ie, we do not differentiate the specific entity classes.
As a result, the parameters of the model can be shared across different domains and classes.
With this in mind, we train the span detection model by exploiting model-agnostic meta-learning~\citep{finn2017model} to promote the learning of the domain-invariant internal representations rather than domain-specific features.
In this way, the meta-learned model is expected to be more sensitive to target-domain support examples, and hence only a few fine-tune steps on these examples can make rapid progress without overfitting.

\subsubsection{Basic Detector}
\paragraph{Model}
In this work, we implement a strong span detector via sequence labeling. We apply the BIOES 
tagging scheme instead of the standard BIO2 to provide more specific and fine-grained boundary information of entity spans.\footnote{We found BIOES to be stronger than BIO for \textit{type-agnostic span detection} as it explicitly encourages the model to learn more specific and fine-grained boundary information. Besides, our entity typing model aims to assign an entity type for each \textit{detected} span, which does not involve any tagging scheme.}
Given an input sequence $\bm{x}=\{x_i\}_{i=1}^L$ with $L$ tokens, we first leverage an encoder $f_{\theta}$ to obtain contextualized representations $\bm{h}=\{h_i\}_{i=1}^L$ for all tokens:
\begin{equation}
    \bm{h} = f_{\theta}(\bm{x}).
    \label{equ:span_encoder}
\end{equation}

With each $h_i$ derived, we then use a linear classification layer to compute the probability distribution of labels that indicate whether the token $x_i$ is inside an entity or not, using a \textit{softmax} function:
\begin{equation}
    p(x_i) = \textrm{softmax}(Wh_i + b),
\end{equation}
where $p(x_i)\in\mathbb{R}^{|C|}$ with $C=\{B, I, O, E, S\}$ being the label set. $\Theta = \{\theta, W, b\}$ are trainable parameters.

\paragraph{Training} Generally, the learning loss \wrt. $\bm{x}$ is modeled as the averaged cross-entropy of the predicted label distribution and the ground-truth one over all tokens.
Following~\citet{wu2020enhanced}, we add a maximum term here to mitigate the problem of insufficient learning for tokens with relatively higher losses, which can be formulated as:
\begin{equation}
\begin{aligned}
    \mathcal{L}(\Theta) &= \frac{1}{L}\sum_{i=1}^L\textrm{CrossEntropy}\left(y_i, p(x_i)\right) \\
    &+ \lambda\max_{i\in\{1, 2, \dots, L\}}\textrm{CrossEntropy}\left(y_i, p(x_i)\right),
\end{aligned}
\label{equ:span_loss}
\end{equation}
where $\lambda \geq 0$ is a weighting factor.

\paragraph{Inference} For inference, we use the learned model to predict the label distribution for each token in a given test case.
We apply the Viterbi algorithm~\citep{forney1973viterbi} for decoding.
It is worthy to note that we do not train a transition matrix here, but simply add constraints to ensure that the predicted label sequence would not violate the BIOES tagging scheme.

\subsubsection{Meta-Learning Procedure}
\label{sec:span_meta}
Here we elaborate on the proposed meta-learning procedure which consists of two phases: meta-training on $\mathcal{E}_{train}$ and meta-testing on $\mathcal{E}_{new}$.
The Appendix \ref{sec:meta} describes the general framework of meta-learning for reference.

\paragraph{Meta-Training} In this phase, we train a mention detection model $\mathcal{M}_{\Theta}$ by repeatedly simulating the \textit{Meta-Testing} phase, where the meta-trained model is fine-tuned with the support set of a novel episode and then tested on the corresponding query set.

Specifically, we first randomly sample an episode $(\mathcal{S}_{train}^{(i)}, \mathcal{Q}_{train}^{(i)}, \mathcal{Y}_{train}^{(i)})$ from $\mathcal{E}_{train}$ and perform \textit{inner-update}:
\begin{equation}
\Theta_i' = U^{n}(\Theta; \alpha, \mathcal{S}_{train}^{(i)}),
\end{equation}
where $ U^{n}$ denotes $n$-step gradient updates with the learning rate $\alpha$ to minimize $\mathcal{L}(\Theta; \mathcal{S}_{train}^{(i)})$, \ie, the loss in Eq.~(\ref{equ:span_loss}) derived from the support set $\mathcal{S}_{train}^{(i)}$.

We then evaluate $\Theta'$ on the query set $\mathcal{Q}_{train}^{(i)}$ and perform \textit{meta-update} by aggregating multiple episodes:
\begin{equation}
    \min_{\Theta}\sum_{i} \mathcal{L}(\Theta_i'; \mathcal{Q}_{train}^{(i)}).
    \label{equ:span_meta_update}
\end{equation}

Since Eq.~(\ref{equ:span_meta_update}) involves the second order derivative, we employ its first-order approximation for computational efficiency:
\begin{equation}
    \Theta \leftarrow \Theta - \beta\sum_{i} \nabla_{\Theta_i'} \mathcal{L}(\Theta_i'; \mathcal{Q}_{train}^{(i)}),
\end{equation}
where $\beta$ denotes the learning rate used in meta-update.

\paragraph{Meta-Testing}
In this phase, we first fine-tune the meta-trained span detection model $\mathcal{M}_{\Theta^*}$ with the loss function defined in Eq.~(\ref{equ:span_loss}) on the support set $\mathcal{S}_{new}$ from a novel episode, and then make predictions for corresponding query examples $\mathcal{Q}_{new}$ with the fine-tuned model $\mathcal{M}_{\Theta'}$ .

\subsection{Entity Typing}
For entity typing, we aim to assign a specific entity class for each span output by the mention detection model.
In the few-shot learning scenario, we take the prototypical networks (ProtoNet)~\citep{snell2017proto} as the backbone for entity typing.
To explore the knowledge brought by support examples from a novel episode, we propose to enhance the ProtoNet with the model-agnostic meta-learning (MAML) algorithm~\citep{finn2017model} for a more representative embedding space, where
text spans from different entity classes are more distinguishable to each other.

\subsubsection{Basic Model: ProtoNet}
\paragraph{Span Representation}
Given an input sequence with $L$ tokens $\bm{x}=\{x_i\}_{i=1}^L$, we use an encoder $g_{\gamma}$ to compute contextual token representations $\bm{h}=\{h_i\}_{i=1}^L$ in the same way as Eq.~(\ref{equ:span_encoder}):
\begin{equation}
    \bm{h} = g_{\gamma}(\bm{x}).
    \label{equ:proto_encoder}
\end{equation}

Assume $x_{[i,j]}$ being the output of the span detection model which starts at $x_i$ and ends at $x_j$, we compute the span representation of $x_{[i,j]}$ by averaging representations of all tokens inside $x_{[i,j]}$:
\begin{equation}
    s_{[i,j]} = \frac{1}{j-i+1}\sum_{k=i}^j h_k.
\end{equation}

\paragraph{Class Prototypes}
Let $S_k=\{x_{[i,j]}\}$ denotes the set of entity spans contained in a given support set $\mathcal{S}$ that belongs to the entity class $y_k\in \mathcal{Y}$,
we compute the prototype $c_k$ for each entity class $y_k$ by averaging span representations of all $x_{[i,j]} \in S_k$:
\begin{equation}
    c_k(\mathcal{S}) = \frac{1}{|S_k|}\sum_{x_{[i,j]}\in S_k}s_{[i,j]}.
    \label{equ:prototype}
\end{equation}

\paragraph{Training} 
Given a training episode denoted as $(\mathcal{S}_{train}, \mathcal{Q}_{train}, \mathcal{Y}_{train})$,
we first utilize the support set $\mathcal{S}_{train}$ to compute prototypes for all entity classes in $\mathcal{Y}_{train}$ via Eq.~(\ref{equ:prototype}).
Then, for each span $x_{[i,j]}$ from the query set $\mathcal{Q}_{train}$, we calculate the probability that $x_{[i,j]}$ belongs to an entity class $y_k\in\mathcal{Y}$ based on the distance between its span representation $s_{[i,j]}$ and the prototype of $y_k$:
\begin{equation}
    p(y_k;x_{[i,j]}) = \frac{\exp\left\{-d\left(c_k(\mathcal{S}_{train}), s_{[i,j]}\right)\right\}}{\sum\limits_{y_i\in\mathcal{Y}}\exp\left\{-d\left(c_i(\mathcal{S}_{train}), s_{[i,j]}\right)\right\}},
    \label{equ:proto_p}
\end{equation}
where $d(\cdot, \cdot)$ denotes the distance function.
Let $y_{[i,j]}\in\mathcal{Y}$ denote the ground-truth entity class \wrt. $x_{[i,j]}$, the parameters of the ProtoNet, \ie, $\gamma$, are trained to minimize the cross-entropy loss:
\begin{equation}
    \mathcal{L}(\gamma) = \sum_{x_{[i,j]}\in\mathcal{Q}_{train}}-\log p(y_{[i,j]}; x_{[i,j]}).
    \label{equ:proto_loss}
\end{equation}

\paragraph{Inference} During inference time, given a novel episode $(\mathcal{S}_{new}, \mathcal{Q}_{new}, \mathcal{Y}_{new})$ for inference,
we first leverage the learned model to compute prototypes for all $y_k \in \mathcal{Y}_{new}$ on $\mathcal{S}_{new}$.
Then, upon the mention detection model, we inference the entity class for each detected entity span $x_{[i,j]}$ in $\mathcal{Q}_{new}$ by taking the label $y_k\in\mathcal{Y}_{new}$ with the highest probability in Eq.~(\ref{equ:proto_p}):
\begin{equation}
    \hat{y}_{[i,j]} = \arg\max_{y_k}p(y_k;x_{[i,j]}).
    \label{equ:proto_inference}
\end{equation}

\subsubsection{MAML Enhanced ProtoNet}
Here, we elaborate on the procedure to integrate the ProtoNet and the model-agnostic meta-learning. 

\paragraph{Meta-Training} Given a randomly sampled episode $(\mathcal{S}_{train}^{(i)}, \mathcal{Q}_{train}^{(i)}, \mathcal{Y}_{train}^{(i)})$ from $\mathcal{E}_{train}$,
for \textit{inner-update}, we first compute prototypes for each entity class in $\mathcal{Y}_{train}$ using $\mathcal{S}_{train}^{(i)}$ via Eq.~(\ref{equ:prototype}), and then take each span $x_{[i,j]} \in \mathcal{S}_{train}^{(i)}$ as the query item in conventional ProtoNet for gradient update:
\begin{equation}
\gamma_i' = U^{n}(\gamma; \alpha, \mathcal{S}_{train}^{(i)}),
\label{equ:proto_inner_update}
\end{equation}
where $ U^{n}$ denotes $n$-step gradient updates with the learning rate $\alpha$ to minimize the cross-entropy loss $\mathcal{L}(\gamma; \mathcal{S}_{train}^{(i)})$ as in Eq.~(\ref{equ:proto_loss}).

As for \textit{meta-update}, we first re-compute prototypes for each entity class in $\mathcal{Y}_{train}^{(i)}$ with $\gamma'$, \ie, the model parameters obtained from \textit{inner-update}.
After that, we perform \textit{meta-update} by evaluating $\gamma'$ on the query set $\mathcal{Q}_{train}^{(i)}$.
We employ the first-order approximation again for computational efficiency.
When aggregating gradients from multiple episodes, it could be formulated as:
\begin{equation}
    \gamma \leftarrow \gamma - \beta\sum_{i} \nabla_{\gamma_i'} \mathcal{L}(\gamma_i'; \mathcal{Q}_{train}^{(i)}),
    \label{equ:proto_meta_update}
\end{equation}

\paragraph{Meta-Testing} Given $(\mathcal{S}_{new}, \mathcal{Q}_{new}, \mathcal{Y}_{new})$, a novel episode unseen during training, conventional ProtoNet directly adopts the meta-trained model to compute prototypes with $\mathcal{S}_{new}$, and then inference on $\mathcal{Q}_{new}$.
Here, we first take the support examples from $\mathcal{S}_{new}$ to fine-tune the meta-learned model $\gamma^*$ for a few steps in a way the same as Eq.~(\ref{equ:proto_inner_update}), however, the loss is computed on $\mathcal{S}_{new}$.
Then, we leverage $\mathcal{S}_{new}$ again to compute prototypes with the fine-tuned model, and further inference the entity class for each detected span in $\mathcal{Q}_{new}$ as in Eq.~(\ref{equ:proto_inference}).

\section{Experiments}
\subsection{Settings}
\subsubsection{Datasets} We conduct experiments to evaluate the proposed approach on two groups of datasets.
\paragraph{Few-NERD~\citep{ding2021nerd}.} It is annotated with a hierarchy of 8 coarse-grained and 66 fine-grained entity types.
Two tasks are considered on this dataset:
i) \textbf{Intra}, where all entities in train/dev/test splits belong to different coarse-grained types.
ii) \textbf{Inter}, where train/dev/test splits may share coarse-grained types while keeping the fine-grained entity types mutually disjoint.
\footnote{\url{https://github.com/thunlp/Few-NERD}}

\paragraph{Cross-Dataset~\citep{hou2020few}.} Four datasets focusing on four domains are used here: CoNLL-2003~\citep{tjong2003introduction} (news), GUM~\citep{zeldes2017gum} (Wiki) , WNUT-2017~\citep{derczynski2017results} (social), and Ontonotes~\citep{pradhan2013towards} (mixed). We take two domains for training, one for validation, and the remaining for test. For fair comparison, we directly use sampled episodes by \citet{hou2020few}.
For more details of these datasets, please refer to the Appendix \ref{sec:dataset}.

\subsubsection{Evaluation}
For evaluation on \textbf{Few-NERD}, we employ \text{episode evaluation} as in \citet{ding2021nerd} and calculate the precision (P), recall (R), and micro F1-score (F1) over all test episodes.
For evaluation on \textbf{Cross-Dataset}, we calculate P, R, F1 within each episode and then average over all episodes as in \citet{hou2020few}.
For all results, we report the mean and standard deviation based on 5 runs with different seeds. 

\subsubsection{Implementation Details}
We implement our approach with PyTorch 1.9.0\footnote{\url{https://pytorch.org/}}. We leverage two separate BERT models for $f_{\theta}$ in Eq.~(\ref{equ:span_encoder}) and $g_{\gamma}$ in Eq.~(\ref{equ:proto_encoder}), respectively. Following previous methods \citep{hou2020few, ding2021nerd}, we use the BERT-base-uncased model \citep{devlin2019bert}.  
The parameters of the embedding layer are frozen during optimization.
We train all models for 1,000 steps and choose the best model with the validation set. We use a batch size of 32, maximum sequence length of 128, and a dropout probability of 0.2.
For the optimizers, we use AdamW~\citep{loshchilov2018decoupled} with a 1\% linearly scheduled warmup.
We perform grid search for other hyper-parameters and select the best settings with the validation set.
For more details, please refer to the Appendix \ref{sec:details}.

\begin{table*}[t]
    \centering
    \setlength{\tabcolsep}{1mm}
    \resizebox{2\columnwidth}{!}{
    \begin{tabular}{lcccccccc}
    \toprule
        \multirow{3}{*}{\textbf{Models}} & \multicolumn{4}{c}{\textbf{Intra}} & \multicolumn{4}{c}{\textbf{Inter}}\\
        \cmidrule(lr){2-5} \cmidrule(lr){6-9}
        & \multicolumn{2}{c}{\textbf{1$\sim$2-shot}} & \multicolumn{2}{c}{\textbf{5$\sim$10-shot}} & \multicolumn{2}{c}{\textbf{1$\sim$2-shot}} & \multicolumn{2}{c}{\textbf{5$\sim$10-shot}}\\
        \cmidrule(lr){2-3}\cmidrule(lr){4-5} \cmidrule(lr){6-7}\cmidrule(lr){8-9}
         & 5 way & 10 way & 5 way & 10 way & 5 way & 10 way & 5 way & 10 way \\
         \cmidrule(lr){1-1}\cmidrule(lr){2-5} \cmidrule(lr){6-9}
         ProtoBERT$^{\dag}$ & 23.45\small\small{\textpm0.92} & 19.76\small{\textpm0.59} & 41.93\small{\textpm0.55} & 34.61\small{\textpm0.59} & 44.44\small{\textpm0.11} & 39.09\small{\textpm0.87} & 58.80\small{\textpm1.42} & 53.97\small{\textpm0.38} \\
         NNShot$^{\dag}$ & 31.01\small{\textpm1.21} & 21.88\small{\textpm0.23} & 35.74\small{\textpm2.36} & 27.67\small{\textpm1.06} & 54.29\small{\textpm0.40} & 46.98\small{\textpm1.96} & 50.56\small{\textpm3.33} & 50.00\small{\textpm0.36} \\
         StructShot$^{\dag}$ & 35.92\small{\textpm0.69} & 25.38\small{\textpm0.84} & 38.83\small{\textpm1.72} & 26.39\small{\textpm2.59} & 57.33\small{\textpm0.53} & 49.46\small{\textpm0.53} & 57.16\small{\textpm2.09} & 49.39\small{\textpm1.77} \\
         CONTAINER \citep{das2021container} & 40.43 & 33.84 & 53.70 & 47.49 & 55.95 & 48.35 & 61.83 & 57.12 \\
         ESD \citep{wang2021enhanced} & 41.44\small{\textpm1.16} & 32.29\small{\textpm1.10} & 50.68\small{\textpm0.94} & 42.92\small{\textpm0.75} & 66.46\small{\textpm0.49} & 59.95\small{\textpm0.69} & \textbf{74.14\small{\textpm0.80}} & 67.91\small{\textpm1.41} \\
         \textbf{Ours} & \textbf{52.04\small{\textpm0.44}} & \textbf{43.50\small{\textpm0.59}} & \textbf{63.23\small{\textpm0.45}} & \textbf{56.84\small{\textpm0.14}} & \textbf{68.77\small{\textpm0.24}} & \textbf{63.26\small{\textpm0.40}} & 71.62\small{\textpm0.16} & \textbf{68.32\small{\textpm0.10}} \\
        \bottomrule
    \end{tabular}
    }
    \caption{F1 scores with standard deviations on Few-NERD for both inter and intra settings. $^{\dag}$ denotes the results reported in \citet{ding2021nerd}.\footnotemark\, The best results are in \textbf{bold}.}
    \label{tab:performance_comparison_fewnerd}
\end{table*}

\begin{table*}[t]
    \centering
    \setlength{\tabcolsep}{1mm}
    \resizebox{2\columnwidth}{!}{
    \begin{tabular}{lcccccccc}
    \toprule
        \multirow{3}{*}{\textbf{Models}} & \multicolumn{4}{c}{\textbf{1-shot}} & \multicolumn{4}{c}{\textbf{5-shot}}\\
        \cmidrule(lr){2-5} \cmidrule(lr){6-9}

         & News & Wiki & Social & Mixed & News & Wiki & Social & Mixed \\
         \cmidrule(lr){1-1}\cmidrule(lr){2-5} \cmidrule(lr){6-9}
        
         TransferBERT$^{\ddag}$ & 4.75\small{\textpm1.42} & 0.57\small{\textpm0.32} & 2.71\small{\textpm0.72} & 3.46\small{\textpm0.54} & 15.36\small{\textpm2.81} & 3.62\small{\textpm0.57} & 11.08\small{\textpm0.57} & 35.49\small{\textpm7.60} \\
         SimBERT$^{\ddag}$ & 19.22\small{\textpm0.00} & 6.91\small{\textpm0.00} & 5.18\small{\textpm0.00} & 13.99\small{\textpm0.00} & 32.01\small{\textpm0.00} & 10.63\small{\textpm0.00} & 8.20\small{\textpm0.00} & 21.14\small{\textpm0.00} \\
         Matching Network$^{\ddag}$ & 19.50\small{\textpm0.35} & 4.73\small{\textpm0.16} & 17.23\small{\textpm2.75} & 15.06\small{\textpm1.61} & 19.85\small{\textpm0.74} & 5.58\small{\textpm0.23} & 6.61\small{\textpm1.75} & 8.08\small{\textpm0.47} \\
         ProtoBERT$^{\ddag}$ & 32.49\small{\textpm2.01} & 3.89\small{\textpm0.24} & 10.68\small{\textpm1.40} & 6.67\small{\textpm0.46} & 50.06\small{\textpm1.57} & 9.54\small{\textpm0.44} & 17.26\small{\textpm2.65} & 13.59\small{\textpm1.61} \\
         L-TapNet+CDT~\citep{hou2020few} & 44.30\small{\textpm3.15} & 12.04\small{\textpm0.65} & 20.80\small{\textpm1.06} & 15.17\small{\textpm1.25} & 45.35\small{\textpm2.67} & 11.65\small{\textpm2.34} & 23.30\small{\textpm2.80} & 20.95\small{\textpm2.81} \\
         \textbf{Ours} & \textbf{46.09\small{\textpm0.44}} & \textbf{17.54\small{\textpm0.98}} & \textbf{25.14\small{\textpm0.24}} & \textbf{34.13\small{\textpm0.92}} & \textbf{58.18\small{\textpm0.87}} & \textbf{31.36\small{\textpm0.91}} & \textbf{31.02\small{\textpm1.28}} & \textbf{45.55\small{\textpm0.90}}\\
        \bottomrule
    \end{tabular}
    }
    \caption{F1 scores with standard deviations on Cross-Dataset. $^{\ddag}$ denotes the results reported in~\citet{hou2020few}. The best results are in \textbf{bold}.}
    \label{tab:performance_comparison_crossdataset}
\end{table*}

\subsection{Main Results}
\paragraph{Baselines} For FewNERD, we compare the proposed approach to ESD~\citep{wang2021enhanced}, CONTAINER~\citep{das2021container}, and methods from \citet{ding2021nerd}, \eg, ProtoBERT, StructShot, \etc. For Corss-Dataset, we compare our method to L-TapNet+CDT~\citep{hou2020few} and other baselines from \citet{hou2020few}, \eg, TransferBERT, Matching Network, \etc.
Please refer to the Appendix \ref{ref:baselines} for more details about baselines.

\footnotetext{To make fair comparison with CONTAINER \citep{das2021container} and ESD \citep{wang2021enhanced}, we use the data from \url{https://cloud.tsinghua.edu.cn/f/8483dc1a34da4a34ab58/?dl=1}, which corresponds to the results reported in \url{https://arxiv.org/pdf/2105.07464v5.pdf}. For results of our approach on data from \url{https://cloud.tsinghua.edu.cn/f/0e38bd108d7b49808cc4/?dl=1}, please refer to our Github.}
\paragraph{Results}
Table~\ref{tab:performance_comparison_fewnerd} and Table~\ref{tab:performance_comparison_crossdataset} report the results of our approach alongside those reported by previous state-of-the-art methods.\footnote{We also provide the intermediate results, \ie, F1-scores of entity span detection in the Appendix \ref{sec:results_span}.}
It can be seen that our proposed method outperforms the prior methods with a large margin, achieving an performance improvement up to 10.60 F1 scores on Few-NERD (\textit{Intra}, 5way 1$\sim$2 shot) and 19.71 F1 scores on Cross-Dataset (\textit{Wiki}, 5-shot), which well demonstrates the effectiveness of the proposed approach.
Table~\ref{tab:performance_comparison_fewnerd} and Table~\ref{tab:performance_comparison_crossdataset} also depict that compared with the results of Few-NERD \textit{Inter}, where the training episodes and test episodes may be constructed with the data from the same domain while still focusing on different fine-grained entity classes, our approach attains more impressive performance in other settings where exists larger transfer gap, \eg, transferring across different coarse entity classes even different datasets built from different domains.
This suggests that our approach is good at dealing with difficult cases, highlighting the necessity of exploring information contained in target-domain support examples and the strong adaptation ability of our approach.

\subsection{Ablation Study}
To validate the contributions of different components in the proposed approach, we introduce the following variants and baselines for ablation study:
\textit{1) Ours w/o MAML}, where we train both the mention detection model and the ProtoNet in a conventional supervised manner and then fine-tune with few-shot examples.
\textit{2) Ours w/o Span Detector}, where we remove the mention detection step and integrate MAML with token-level prototypical networks.
\textit{3) Ours w/o Span Detector w/o MAML}, where we further eliminate the meta-learning procedure from \textit{Ours w/o Span Detector}, and thus becomes the conventional token-level prototypical networks.
\textit{4) Ours w/o ProtoNet}, where we directly apply the original MAML algorithm to train a BERT-based tagger for few-shot NER.
\begin{table}[htb]
    \centering
    \resizebox{\columnwidth}{!}{
    \begin{tabular}{lcc}
    \toprule
        & \textbf{Intra} & \textbf{Inter} \\
    \midrule
        \textbf{Ours} & \textbf{52.04} & \textbf{68.77} \\
        1) Ours w/o MAML & 48.76 & 64.44\\
        2) Ours w/o Span Detector  & 36.06 & 53.56\\
        3) Ours w/o Span Detector w/o MAML & 23.45  & 44.44 \\
        4) Ours w/o ProtoNet & 21.20 & 45.71\\
    \bottomrule
    \end{tabular}
    }
    \caption{Ablation study: F1 scores on Few-NERD 5-way 1$\sim$2-shot are reported.}
    \label{tab:ablation}
\end{table}

\begin{figure*}[t]
    \centering
    \includegraphics[width=2\columnwidth]{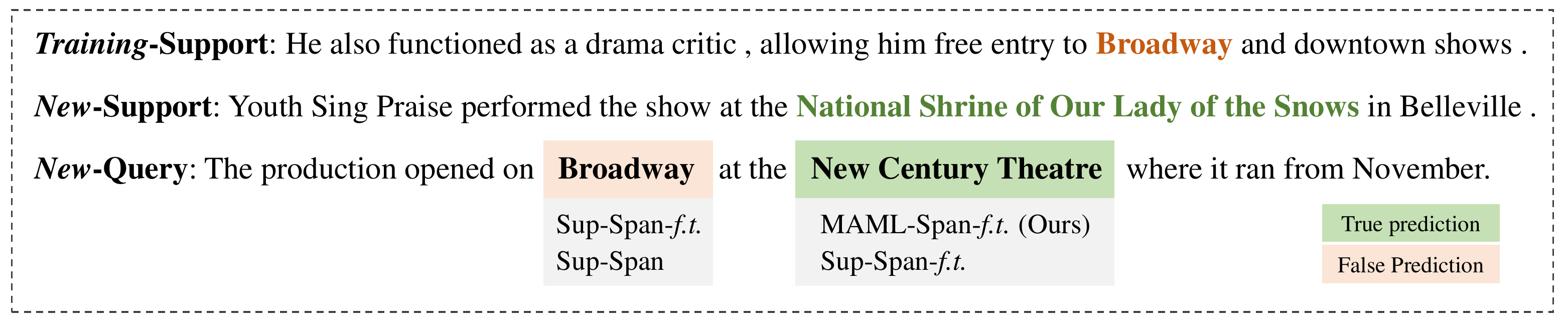}
    \caption{Case study of span detection. \textbf{Sup-Span}: train a span detector in the fully supervised manner on available data from all training episodes, and then directly use it for span detection. \textbf{Sup-Span-f.t.}: further fine-tune the model learned by \textit{Sup-Span} as in the proposed approach.}
    \label{fig:case_study}
\end{figure*}

Table~\ref{tab:ablation} highlights the contributions of each component in our proposed approach.
Generally speaking, removing any of them will generally lead to a performance drop.
Moreover, we can draw some in-depth observations as follows.
1) \textit{Ours} outperforms \textit{Ours w/o MAML} and \textit{Ours w/o Span Detector} outperforms \textit{Ours w/o Span Detector w/o MAML} indicate that exploring information contained in support examples with the proposed meta-learning procedure does bring performance gain for few-shot transfer.
2) \textit{Ours} outperforms \textit{Ours w/o Span Detector} and \textit{Ours w/o MAML} outperforms \textit{Ours w/o Span Detector w/o MAML} demonstrate the essentiality of the decomposed framework (\ie, mention detection and entity typing) to mitigate the problem of noisy prototype for non-entities.
3) Though MAML plays an important role in learning from few-shot support examples, 
\textit{Ours w/o ProtoNet}, which requires the model to adapt the up-most classification layer without sharing knowledge with training episodes leads to unsatisfactory results, verifying the reasonableness and the effectiveness of our decomposed meta-learning procedure.


\paragraph{How does MAML promote the span detector?}
To bring up insights on how MAML promotes the span detector, here we introduce two baselines and compare them to our approach by case study. 
As shown in Figure~\ref{fig:case_study},  given a query sentence from a novel episode, \textit{Sup-Span} only predicts a false positive span ``Broadway'' while missing the golden span ``New Century Theatre''.
Note that ``Broadway'' appears in training corpus as an entity span, indicating that the span detector trained in a fully supervised manner performs well on seen entity spans, but struggles to detect un-seen entity spans.
Figure~\ref{fig:case_study} also shows that both our method and \textit{Sup-Span-f.t.} can successfully detect ``New Century Theatre''.
However, \textit{Sup-Span-f.t.} still outputs ``Broadway'' while our method can produce more accurate predictions.
This shows that though fine-tuning can benefit full supervised model on new entity classes to some extend, it may bias too much to the training data.

\begin{figure}[t]
  \centering
  \subfloat[Intra]{
    \label{sfig:intra-finetune-steps}
    \includegraphics[height=2.44cm]{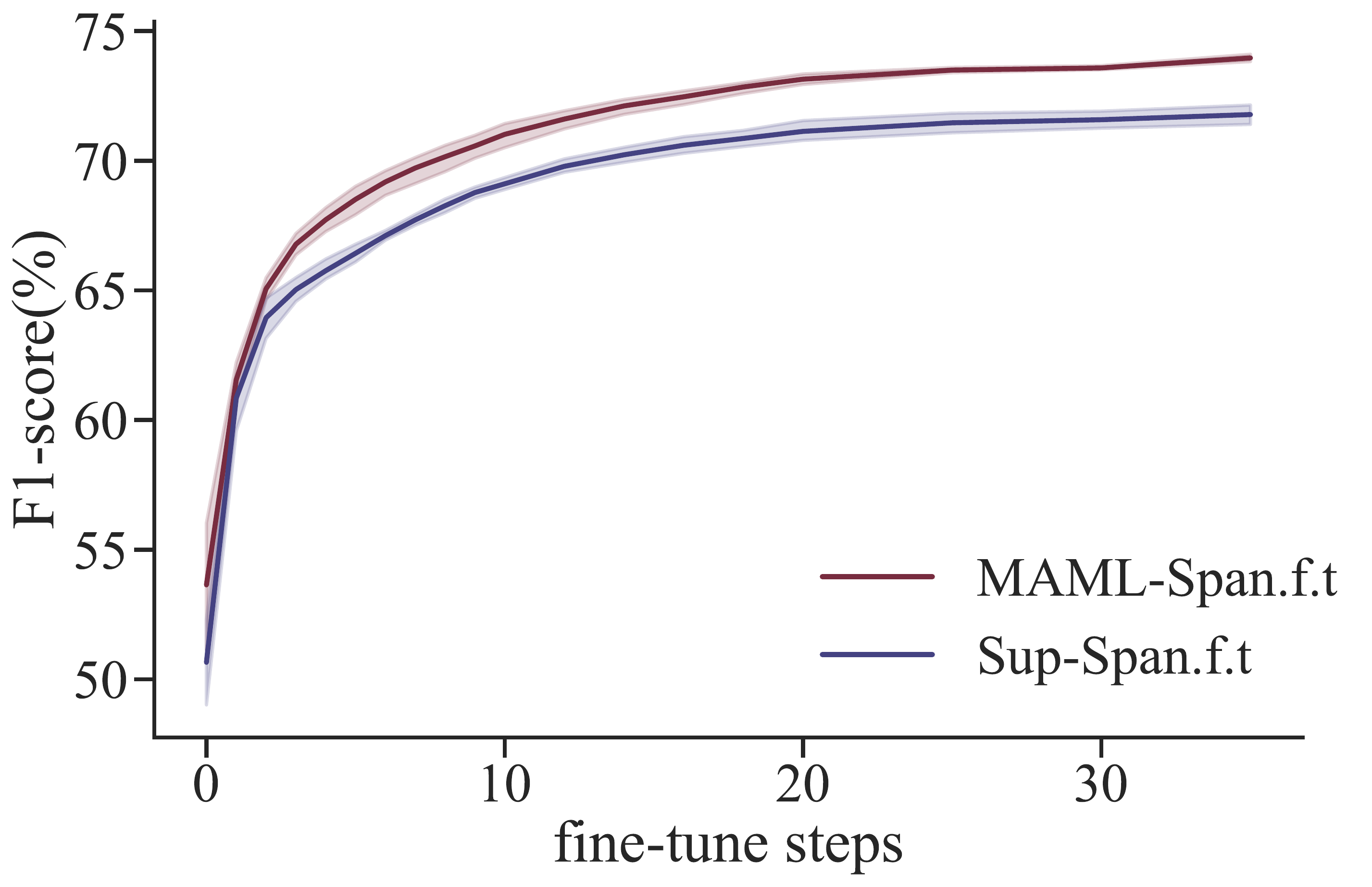}}
  \subfloat[Inter]{
    \label{sfig:inter-finetune-steps}
    \includegraphics[height=2.44cm]{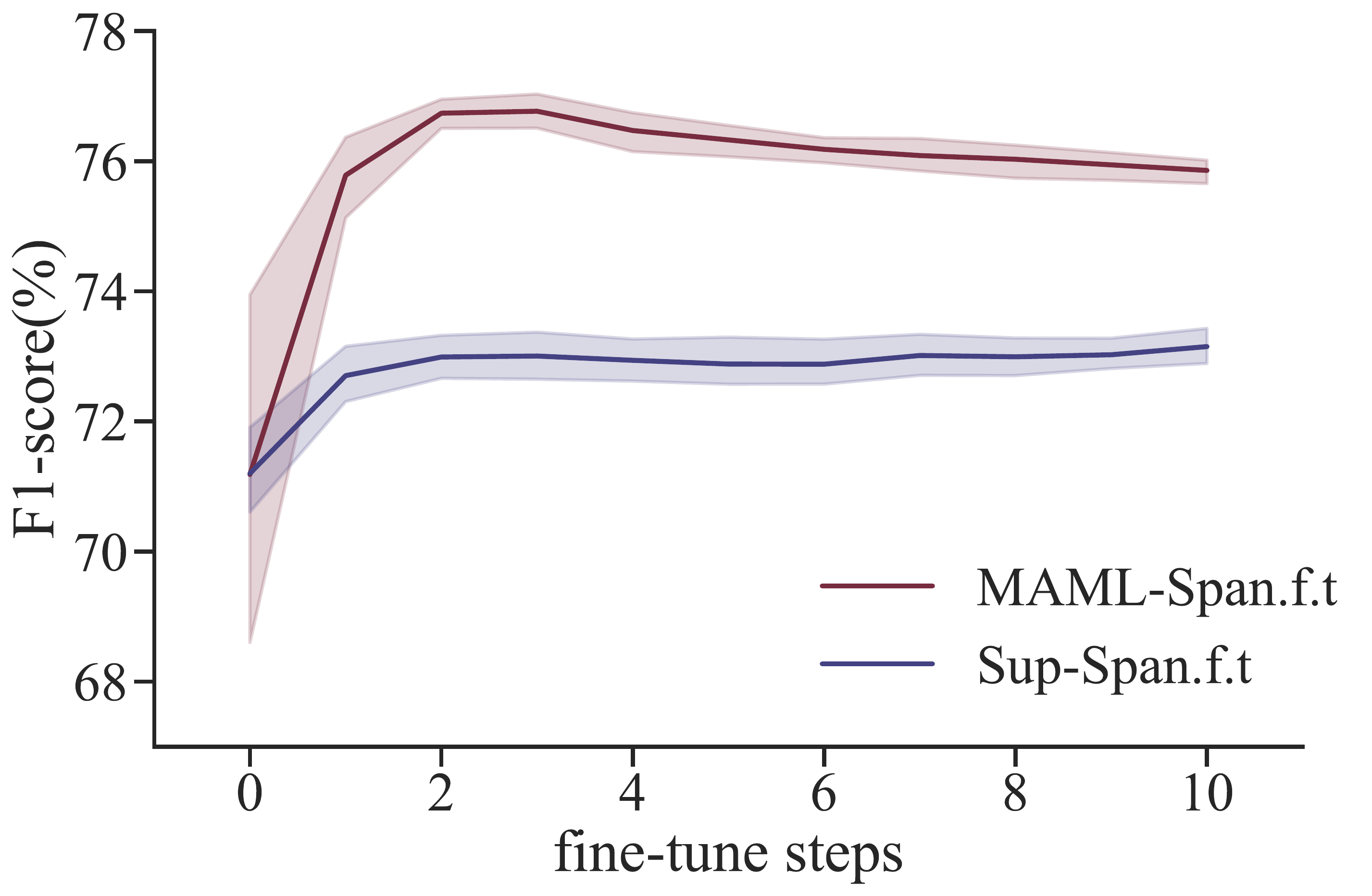}}
  \caption{F1 scores of differently trained span detectors \wrt. fine-tune steps on Few-NERD 5-way 1$\sim$2-shot test set. The light-colored area indicates the range of results obtained from multiple random seeds.
  }
  \label{fig:analysis_finetune_steps}
\end{figure}
We further investigate how performances of aforementioned span detectors vary with different fine-tune steps.
As shown in Figure~\ref{fig:analysis_finetune_steps},
our model (\textit{MAML-Span-f.t.}) consistently outperforms \textit{Sup-Span-f.t.}, suggesting that the proposed meta-learning procedure could better leverage support examples from novel episodes and meanwhile, help the model adapt to new episodes more effectively.

\begin{table}[t]
    \centering
    \resizebox{\columnwidth}{!}{
    \begin{tabular}{lcc}
    \toprule
         & \textbf{Intra} & \textbf{Inter} \\
    \midrule
        \textbf{Ours} (w/ MAML-ProtoNet) & \textbf{52.04} & \textbf{68.77} \\
        Ours w/ ProtoNet & 50.53 & 67.79\\
    \bottomrule
    \end{tabular}
    }
    \caption{Analysis on entity typing under Few-NERD 5-way 1$\sim$2-shot setting. F1 scores are reported. \textbf{Ours w/ ProtoNet}: built upon the same span detection model as \textit{Ours}, directly leverage ProtoNet for inference.}
    \label{tab:analysis_proto}
\end{table}
\begin{figure}[t]
  \centering
  \subfloat[Proto]{
    \label{sfig:proto-span-embedding}
    \includegraphics[height=2.44cm]{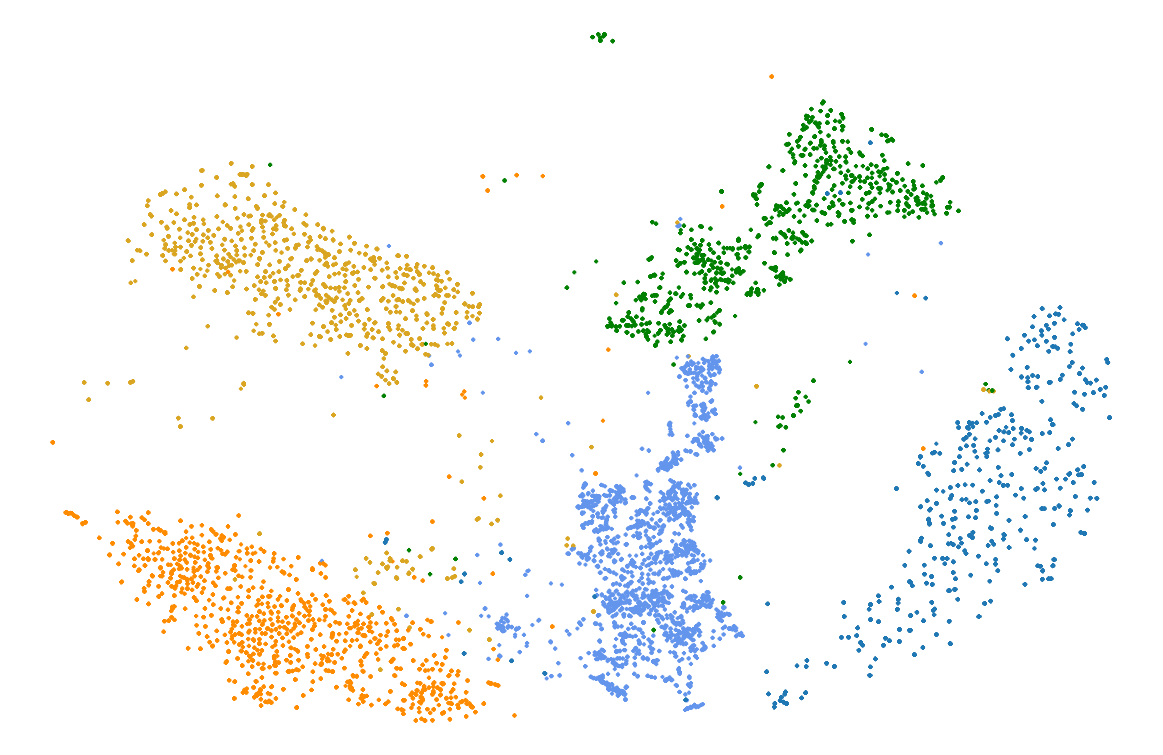}}
  \subfloat[MAML-Proto]{
    \label{sfig:maml-proto-span-embedding}
    \includegraphics[height=2.44cm]{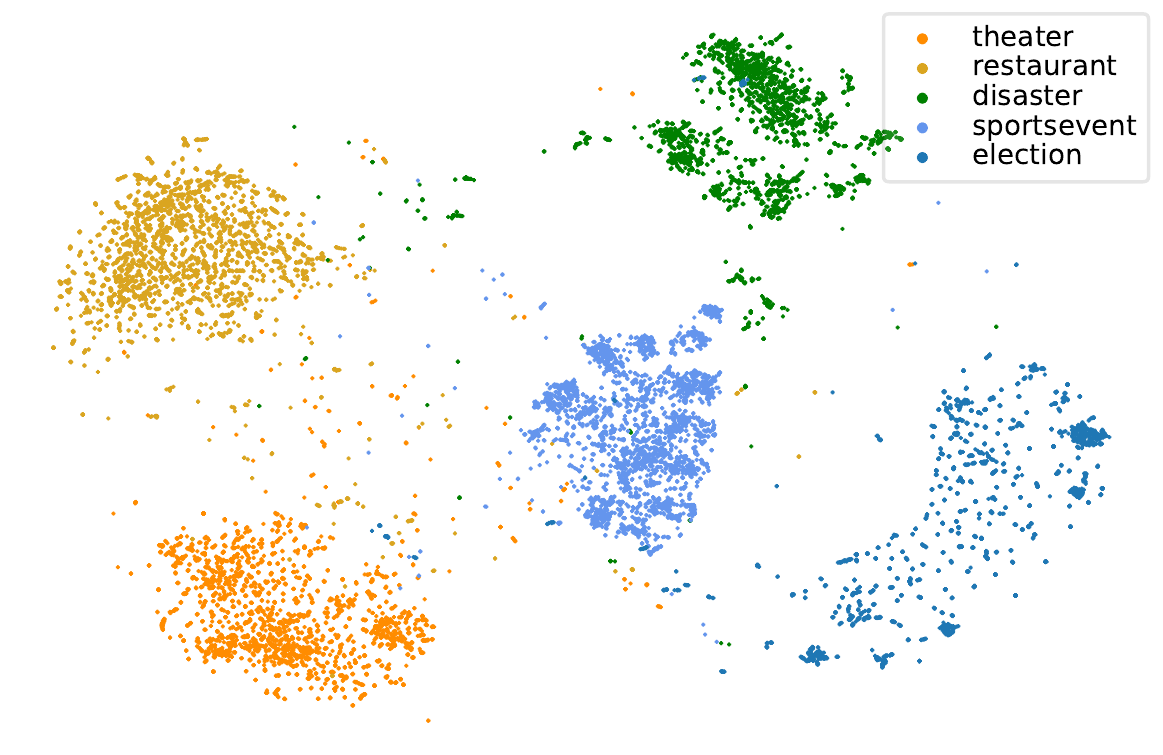}}
  \caption{t-SNE visualization of span representations for entity typing on  Few-NERD \textit{Intra}, 5-way 5$\sim$10-shot dev set. The representations are obtained from BERT trained with ProtoNet, and our MAML enhanced ProtoNet respectively.}
  \label{fig:analysis_proto}
\end{figure}
\paragraph{How does MAML enhance the ProtoNet?} 
We first compare the proposed MAML-Proto to the conventional ProtoNet based on the same span detector proposed in this paper.
Table~\ref{tab:analysis_proto} shows that our MAML-ProtoNet achieves superior performance than the conventional ProtoNet, which verifies the effectiveness of leveraging the support examples to refine the learned embedding space at test time.
To further analyze how MAML adjusts the representation space of entity spans and prototypes, we utilize  \textit{t-SNE}~\citep{van2008tsne} to reduce the dimension of span representations obtained from ProtoNet and MAML-ProtoNet for entity typing, the visualization is shown in Figure~\ref{fig:analysis_proto}.
We can see that MAML enhanced Proto can cluster span representations of the same entity class while dispersing span representations of different entity classes .
Therefore, compared with ProtoNet, it is easier for the proposed MAML-ProtoNet to assign an entity class for a query span by measuring similarities between its representation and the prototype of each entity class.

\section{Related Work}
\paragraph{Neural NER}
Modern NER systems usually formulate the NER task as a sequence labeling problem and tackle it by implementing deep neural networks and a token-level classification layer with a conditional random field \citep[CRF]{Lafferty2001conditional} layer on top~\citep{ma2016end, chiu2016named, liu2019towards, devlin2019bert}.
Alternative approaches for NER are also proposed to handle the problem based on span classification \citep{ouchi-2020-instance, fu-etal-2021-spanner}, machine reading comprehension \citep{li2020unified}, and sequence generation \citep{yan-etal-2021-ner-generative}.

\paragraph{Few-Shot Learning and Meta-Learning} 
Recently, few-shot learning
has received increasing attention in the NLP community \citep{han2018fewrel, geng-2019-induction, chen-2019-metalink, brown2020gpt3, schick-schutze-2021-just, gao2021}.
and meta-learning has become a popular paradigm for few-shot settings.
Typical meta-learning approaches can be divided into three categories: black-box adaption based methods \citep{santoro-2016-memoryaug}, optimization based methods \citep{finn2017model}, and metric learning based methods\citep{vinyal-2016-matchnet, snell2017proto}. Our work takes advantages of two popular meta-learning approaches, \ie, prototypical network \citep{snell2017proto} and MAML \citep{finn2017model}.
The most related work of this paper is \citet{trian2020metadata}, which similarly implements MAML updates over prototypical networks for few-shot image classification.
\paragraph{Few-Shot NER}
Studies on few-shot NER typically adopt metric learning based approaches at either token-level \citep{fritzler2019few, hou2020few, yang2020simple, tong2021fewother} or span-level \citep{yu2021fewShot, wang2021enhanced}. 
\citet{athiwaratkun2020augmented} and \citet{cui2021template} also propose to address the problem via sequence generation and adapt the model to a new domain within the conventional transfer learning paradigm (training plus finetuning).
Differently, \citet{wang2021learning} propose to decompose the problem into span detection and entity type classification to better leverage type description.
They exploit a traditional span-based classifier to detect entity spans and leverage class descriptions to learn representations for each entity class. 
When adapting the model to new domains in the few-shot setting, they directly fine-tune the model with the support examples. 
In this paper, we propose a decomposed meta-learning based method to handle few-shot span detection and few-shot entity typing sequentially for few-shot NER.
The contribution and novelty of our work lie in that:
i) Previous work transfers the metric-learning based model learned in source domains to a novel target domain either without any parameter updates \citep{hou2020few, wang2021enhanced} or by simply applying conventional fine-tuning \citep{cui2021template, das2021container, wang2021learning}, 
while we introduce the model-agnostic meta-learning and integrate it with the prevalent prototypical networks to leverage the information contained in support examples more effectively. 
ii) Existing studies depend on one \citep{hou2020few} or multiple prototypes \citep{tong2021fewother, wang2021enhanced} to represent text spans of non-entities (``\texttt{O}'') for class inference,
while we avoid this problem by only locating named entities during span detection.
Moreover, meta-learning has also been exploited in a few recent studies \citep{li2020metaner,lichy2021meta} for few-shot NER.
However, our work substantially differs from them in that we proposed a decomposed meta-learning procedure to separately optimize the span detection model and the entity typing model.

\section{Conclusion}
This paper presents a decomposed meta-learning method for few-shot NER problem, \ie,
sequentially tackle few-shot span-detection and few-shot entity typing using meta-learning.
We formulate the few-shot span detection as a sequence labeling problem and employ MAML to learn a good parameter initialization, which enables the model to fast adapt to novel entity classes by fully exploring information contained in support examples.
For few-shot entity typing, we propose MAML-ProtoNet, which can find a better embedding space than conventional ProtoNet to represent entity spans from different classes more distinguishably, thus making more accurate predictions.
Extensive experiments on various benchmarks show that our approach achieves superior performance over prior methods.
\bibliographystyle{acl_natbib}

\begin{thebibliography}{45}
\expandafter\ifx\csname natexlab\endcsname\relax\def\natexlab#1{#1}\fi

\bibitem[{Athiwaratkun et~al.(2020)Athiwaratkun, Nogueira~dos Santos, Krone,
  and Xiang}]{athiwaratkun2020augmented}
Ben Athiwaratkun, Cicero Nogueira~dos Santos, Jason Krone, and Bing Xiang.
  2020.
\newblock \href {https://doi.org/10.18653/v1/2020.emnlp-main.27} {Augmented
  natural language for generative sequence labeling}.
\newblock In \emph{Proceedings of the 2020 Conference on Empirical Methods in
  Natural Language Processing (EMNLP)}, pages 375--385, Online. Association for
  Computational Linguistics.

\bibitem[{Brown et~al.(2020)Brown, Mann, Ryder, Subbiah, Kaplan, Dhariwal,
  Neelakantan, Shyam, Sastry, Askell, Agarwal, Herbert{-}Voss, Krueger,
  Henighan, Child, Ramesh, Ziegler, Wu, Winter, Hesse, Chen, Sigler, Litwin,
  Gray, Chess, Clark, Berner, McCandlish, Radford, Sutskever, and
  Amodei}]{brown2020gpt3}
Tom~B. Brown, Benjamin Mann, Nick Ryder, Melanie Subbiah, Jared Kaplan,
  Prafulla Dhariwal, Arvind Neelakantan, Pranav Shyam, Girish Sastry, Amanda
  Askell, Sandhini Agarwal, Ariel Herbert{-}Voss, Gretchen Krueger, Tom
  Henighan, Rewon Child, Aditya Ramesh, Daniel~M. Ziegler, Jeffrey Wu, Clemens
  Winter, Christopher Hesse, Mark Chen, Eric Sigler, Mateusz Litwin, Scott
  Gray, Benjamin Chess, Jack Clark, Christopher Berner, Sam McCandlish, Alec
  Radford, Ilya Sutskever, and Dario Amodei. 2020.
\newblock \href
  {https://proceedings.neurips.cc/paper/2020/hash/1457c0d6bfcb4967418bfb8ac142f64a-Abstract.html}
  {Language models are few-shot learners}.
\newblock In \emph{Advances in Neural Information Processing Systems 33: Annual
  Conference on Neural Information Processing Systems 2020, NeurIPS 2020,
  December 6-12, 2020, virtual}.

\bibitem[{Chen et~al.(2019)Chen, Zhang, Zhang, Chen, and
  Chen}]{chen-2019-metalink}
Mingyang Chen, Wen Zhang, Wei Zhang, Qiang Chen, and Huajun Chen. 2019.
\newblock \href {https://doi.org/10.18653/v1/D19-1431} {Meta relational
  learning for few-shot link prediction in knowledge graphs}.
\newblock In \emph{Proceedings of the 2019 Conference on Empirical Methods in
  Natural Language Processing and the 9th International Joint Conference on
  Natural Language Processing (EMNLP-IJCNLP)}, pages 4217--4226, Hong Kong,
  China. Association for Computational Linguistics.

\bibitem[{Chiu and Nichols(2016)}]{chiu2016named}
Jason~P.C. Chiu and Eric Nichols. 2016.
\newblock \href {https://doi.org/10.1162/tacl_a_00104} {Named entity
  recognition with bidirectional {LSTM}-{CNN}s}.
\newblock \emph{Transactions of the Association for Computational Linguistics},
  4:357--370.

\bibitem[{Cui et~al.(2021)Cui, Wu, Liu, Yang, and Zhang}]{cui2021template}
Leyang Cui, Yu~Wu, Jian Liu, Sen Yang, and Yue Zhang. 2021.
\newblock \href {https://doi.org/10.18653/v1/2021.findings-acl.161}
  {Template-based named entity recognition using {BART}}.
\newblock In \emph{Findings of the Association for Computational Linguistics:
  ACL-IJCNLP 2021}, pages 1835--1845, Online. Association for Computational
  Linguistics.

\bibitem[{Das et~al.(2021)Das, Katiyar, Passonneau, and
  Zhang}]{das2021container}
Sarkar Snigdha~Sarathi Das, Arzoo Katiyar, Rebecca~J Passonneau, and Rui Zhang.
  2021.
\newblock \href {https://arxiv.org/abs/2109.07589} {Container: Few-shot named
  entity recognition via contrastive learning}.
\newblock \emph{ArXiv preprint}, abs/2109.07589.

\bibitem[{de~Lichy et~al.(2021)de~Lichy, Glaude, and Campbell}]{lichy2021meta}
Cyprien de~Lichy, Hadrien Glaude, and William Campbell. 2021.
\newblock \href {https://doi.org/10.18653/v1/2021.metanlp-1.6} {Meta-learning
  for few-shot named entity recognition}.
\newblock In \emph{Proceedings of the 1st Workshop on Meta Learning and Its
  Applications to Natural Language Processing}, pages 44--58, Online.
  Association for Computational Linguistics.

\bibitem[{Derczynski et~al.(2017)Derczynski, Nichols, van Erp, and
  Limsopatham}]{derczynski2017results}
Leon Derczynski, Eric Nichols, Marieke van Erp, and Nut Limsopatham. 2017.
\newblock \href {https://doi.org/10.18653/v1/W17-4418} {Results of the
  {WNUT}2017 shared task on novel and emerging entity recognition}.
\newblock In \emph{Proceedings of the 3rd Workshop on Noisy User-generated
  Text}, pages 140--147, Copenhagen, Denmark. Association for Computational
  Linguistics.

\bibitem[{Devlin et~al.(2019)Devlin, Chang, Lee, and
  Toutanova}]{devlin2019bert}
Jacob Devlin, Ming-Wei Chang, Kenton Lee, and Kristina Toutanova. 2019.
\newblock \href {https://doi.org/10.18653/v1/N19-1423} {{BERT}: Pre-training of
  deep bidirectional transformers for language understanding}.
\newblock In \emph{Proceedings of the 2019 Conference of the North {A}merican
  Chapter of the Association for Computational Linguistics: Human Language
  Technologies, Volume 1 (Long and Short Papers)}, pages 4171--4186,
  Minneapolis, Minnesota. Association for Computational Linguistics.

\bibitem[{Ding et~al.(2021)Ding, Xu, Chen, Wang, Han, Xie, Zheng, and
  Liu}]{ding2021nerd}
Ning Ding, Guangwei Xu, Yulin Chen, Xiaobin Wang, Xu~Han, Pengjun Xie, Haitao
  Zheng, and Zhiyuan Liu. 2021.
\newblock \href {https://doi.org/10.18653/v1/2021.acl-long.248} {Few-{NERD}: A
  few-shot named entity recognition dataset}.
\newblock In \emph{Proceedings of the 59th Annual Meeting of the Association
  for Computational Linguistics and the 11th International Joint Conference on
  Natural Language Processing (Volume 1: Long Papers)}, pages 3198--3213,
  Online. Association for Computational Linguistics.

\bibitem[{Finn et~al.(2017)Finn, Abbeel, and Levine}]{finn2017model}
Chelsea Finn, Pieter Abbeel, and Sergey Levine. 2017.
\newblock \href {http://proceedings.mlr.press/v70/finn17a.html} {Model-agnostic
  meta-learning for fast adaptation of deep networks}.
\newblock In \emph{Proceedings of the 34th International Conference on Machine
  Learning, {ICML} 2017, Sydney, NSW, Australia, 6-11 August 2017}, volume~70
  of \emph{Proceedings of Machine Learning Research}, pages 1126--1135. {PMLR}.

\bibitem[{Forney(1973)}]{forney1973viterbi}
G~David Forney. 1973.
\newblock The viterbi algorithm.
\newblock \emph{Proceedings of the IEEE}, 61(3):268--278.

\bibitem[{Fritzler et~al.(2019)Fritzler, Logacheva, and
  Kretov}]{fritzler2019few}
Alexander Fritzler, Varvara Logacheva, and Maksim Kretov. 2019.
\newblock Few-shot classification in named entity recognition task.
\newblock In \emph{Proceedings of the 34th ACM/SIGAPP Symposium on Applied
  Computing}, pages 993--1000.

\bibitem[{Fu et~al.(2021)Fu, Huang, and Liu}]{fu-etal-2021-spanner}
Jinlan Fu, Xuanjing Huang, and Pengfei Liu. 2021.
\newblock \href {https://doi.org/10.18653/v1/2021.acl-long.558} {{S}pan{NER}:
  Named entity re-/recognition as span prediction}.
\newblock In \emph{Proceedings of the 59th Annual Meeting of the Association
  for Computational Linguistics and the 11th International Joint Conference on
  Natural Language Processing (Volume 1: Long Papers)}, pages 7183--7195,
  Online. Association for Computational Linguistics.

\bibitem[{Gao et~al.(2021)Gao, Fisch, and Chen}]{gao2021}
Tianyu Gao, Adam Fisch, and Danqi Chen. 2021.
\newblock \href {https://doi.org/10.18653/v1/2021.acl-long.295} {Making
  pre-trained language models better few-shot learners}.
\newblock In \emph{Proceedings of the 59th Annual Meeting of the Association
  for Computational Linguistics and the 11th International Joint Conference on
  Natural Language Processing (Volume 1: Long Papers)}, pages 3816--3830,
  Online. Association for Computational Linguistics.

\bibitem[{Geng et~al.(2019)Geng, Li, Li, Zhu, Jian, and
  Sun}]{geng-2019-induction}
Ruiying Geng, Binhua Li, Yongbin Li, Xiaodan Zhu, Ping Jian, and Jian Sun.
  2019.
\newblock \href {https://doi.org/10.18653/v1/D19-1403} {Induction networks for
  few-shot text classification}.
\newblock In \emph{Proceedings of the 2019 Conference on Empirical Methods in
  Natural Language Processing and the 9th International Joint Conference on
  Natural Language Processing (EMNLP-IJCNLP)}, pages 3904--3913, Hong Kong,
  China. Association for Computational Linguistics.

\bibitem[{Han et~al.(2018)Han, Zhu, Yu, Wang, Yao, Liu, and
  Sun}]{han2018fewrel}
Xu~Han, Hao Zhu, Pengfei Yu, Ziyun Wang, Yuan Yao, Zhiyuan Liu, and Maosong
  Sun. 2018.
\newblock \href {https://doi.org/10.18653/v1/D18-1514} {{F}ew{R}el: A
  large-scale supervised few-shot relation classification dataset with
  state-of-the-art evaluation}.
\newblock In \emph{Proceedings of the 2018 Conference on Empirical Methods in
  Natural Language Processing}, pages 4803--4809, Brussels, Belgium.
  Association for Computational Linguistics.

\bibitem[{Hou et~al.(2020)Hou, Che, Lai, Zhou, Liu, Liu, and Liu}]{hou2020few}
Yutai Hou, Wanxiang Che, Yongkui Lai, Zhihan Zhou, Yijia Liu, Han Liu, and Ting
  Liu. 2020.
\newblock \href {https://doi.org/10.18653/v1/2020.acl-main.128} {Few-shot slot
  tagging with collapsed dependency transfer and label-enhanced task-adaptive
  projection network}.
\newblock In \emph{Proceedings of the 58th Annual Meeting of the Association
  for Computational Linguistics}, pages 1381--1393, Online. Association for
  Computational Linguistics.

\bibitem[{Lafferty et~al.(2001)Lafferty, McCallum, and
  Pereira}]{Lafferty2001conditional}
John~D. Lafferty, Andrew McCallum, and Fernando C.~N. Pereira. 2001.
\newblock Conditional random fields: Probabilistic models for segmenting and
  labeling sequence data.
\newblock In \emph{Proceedings of the Eighteenth International Conference on
  Machine Learning {(ICML} 2001), Williams College, Williamstown, MA, USA, June
  28 - July 1, 2001}, pages 282--289. Morgan Kaufmann.

\bibitem[{Lample et~al.(2016)Lample, Ballesteros, Subramanian, Kawakami, and
  Dyer}]{lample2016neural}
Guillaume Lample, Miguel Ballesteros, Sandeep Subramanian, Kazuya Kawakami, and
  Chris Dyer. 2016.
\newblock \href {https://doi.org/10.18653/v1/N16-1030} {Neural architectures
  for named entity recognition}.
\newblock In \emph{Proceedings of the 2016 Conference of the North {A}merican
  Chapter of the Association for Computational Linguistics: Human Language
  Technologies}, pages 260--270, San Diego, California. Association for
  Computational Linguistics.

\bibitem[{Li et~al.(2020{\natexlab{a}})Li, Chiu, Feng, and
  Wang}]{li2020metaner}
J.~Li, B.~Chiu, S.~Feng, and H.~Wang. 2020{\natexlab{a}}.
\newblock \href {https://doi.org/10.1109/TKDE.2020.3038670} {Few-shot named
  entity recognition via meta-learning}.
\newblock \emph{IEEE Transactions on Knowledge \& Data Engineering}.

\bibitem[{Li et~al.(2020{\natexlab{b}})Li, Feng, Meng, Han, Wu, and
  Li}]{li2020unified}
Xiaoya Li, Jingrong Feng, Yuxian Meng, Qinghong Han, Fei Wu, and Jiwei Li.
  2020{\natexlab{b}}.
\newblock \href {https://doi.org/10.18653/v1/2020.acl-main.519} {A unified
  {MRC} framework for named entity recognition}.
\newblock In \emph{Proceedings of the 58th Annual Meeting of the Association
  for Computational Linguistics}, pages 5849--5859, Online. Association for
  Computational Linguistics.

\bibitem[{Liu et~al.(2019)Liu, Yao, and Lin}]{liu2019towards}
Tianyu Liu, Jin-Ge Yao, and Chin-Yew Lin. 2019.
\newblock \href {https://doi.org/10.18653/v1/P19-1524} {Towards improving
  neural named entity recognition with gazetteers}.
\newblock In \emph{Proceedings of the 57th Annual Meeting of the Association
  for Computational Linguistics}, pages 5301--5307, Florence, Italy.
  Association for Computational Linguistics.

\bibitem[{Loshchilov and Hutter(2019)}]{loshchilov2018decoupled}
Ilya Loshchilov and Frank Hutter. 2019.
\newblock \href {https://openreview.net/forum?id=Bkg6RiCqY7} {Decoupled weight
  decay regularization}.
\newblock In \emph{7th International Conference on Learning Representations,
  {ICLR} 2019, New Orleans, LA, USA, May 6-9, 2019}. OpenReview.net.

\bibitem[{Ma and Hovy(2016)}]{ma2016end}
Xuezhe Ma and Eduard Hovy. 2016.
\newblock \href {https://doi.org/10.18653/v1/P16-1101} {End-to-end sequence
  labeling via bi-directional {LSTM}-{CNN}s-{CRF}}.
\newblock In \emph{Proceedings of the 54th Annual Meeting of the Association
  for Computational Linguistics (Volume 1: Long Papers)}, pages 1064--1074,
  Berlin, Germany. Association for Computational Linguistics.

\bibitem[{Ouchi et~al.(2020)Ouchi, Suzuki, Kobayashi, Yokoi, Kuribayashi,
  Konno, and Inui}]{ouchi-2020-instance}
Hiroki Ouchi, Jun Suzuki, Sosuke Kobayashi, Sho Yokoi, Tatsuki Kuribayashi,
  Ryuto Konno, and Kentaro Inui. 2020.
\newblock \href {https://doi.org/10.18653/v1/2020.acl-main.575} {Instance-based
  learning of span representations: A case study through named entity
  recognition}.
\newblock In \emph{Proceedings of the 58th Annual Meeting of the Association
  for Computational Linguistics}, pages 6452--6459, Online. Association for
  Computational Linguistics.

\bibitem[{Peters et~al.(2017)Peters, Ammar, Bhagavatula, and
  Power}]{peters2017semi}
Matthew~E. Peters, Waleed Ammar, Chandra Bhagavatula, and Russell Power. 2017.
\newblock \href {https://doi.org/10.18653/v1/P17-1161} {Semi-supervised
  sequence tagging with bidirectional language models}.
\newblock In \emph{Proceedings of the 55th Annual Meeting of the Association
  for Computational Linguistics (Volume 1: Long Papers)}, pages 1756--1765,
  Vancouver, Canada. Association for Computational Linguistics.

\bibitem[{Pradhan et~al.(2013)Pradhan, Moschitti, Xue, Ng, Bj{\"o}rkelund,
  Uryupina, Zhang, and Zhong}]{pradhan2013towards}
Sameer Pradhan, Alessandro Moschitti, Nianwen Xue, Hwee~Tou Ng, Anders
  Bj{\"o}rkelund, Olga Uryupina, Yuchen Zhang, and Zhi Zhong. 2013.
\newblock \href {https://aclanthology.org/W13-3516} {Towards robust linguistic
  analysis using {O}nto{N}otes}.
\newblock In \emph{Proceedings of the Seventeenth Conference on Computational
  Natural Language Learning}, pages 143--152, Sofia, Bulgaria. Association for
  Computational Linguistics.

\bibitem[{Santoro et~al.(2016)Santoro, Bartunov, Botvinick, Wierstra, and
  Lillicrap}]{santoro-2016-memoryaug}
Adam Santoro, Sergey Bartunov, Matthew Botvinick, Daan Wierstra, and Timothy~P.
  Lillicrap. 2016.
\newblock \href {http://proceedings.mlr.press/v48/santoro16.html}
  {Meta-learning with memory-augmented neural networks}.
\newblock In \emph{Proceedings of the 33nd International Conference on Machine
  Learning, {ICML} 2016, New York City, NY, USA, June 19-24, 2016}, volume~48
  of \emph{{JMLR} Workshop and Conference Proceedings}, pages 1842--1850.
  JMLR.org.

\bibitem[{Schick and Sch{\"u}tze(2021)}]{schick-schutze-2021-just}
Timo Schick and Hinrich Sch{\"u}tze. 2021.
\newblock \href {https://doi.org/10.18653/v1/2021.naacl-main.185} {It{'}s not
  just size that matters: Small language models are also few-shot learners}.
\newblock In \emph{Proceedings of the 2021 Conference of the North American
  Chapter of the Association for Computational Linguistics: Human Language
  Technologies}, pages 2339--2352, Online. Association for Computational
  Linguistics.

\bibitem[{Snell et~al.(2017)Snell, Swersky, and Zemel}]{snell2017proto}
Jake Snell, Kevin Swersky, and Richard~S. Zemel. 2017.
\newblock \href
  {https://proceedings.neurips.cc/paper/2017/hash/cb8da6767461f2812ae4290eac7cbc42-Abstract.html}
  {Prototypical networks for few-shot learning}.
\newblock In \emph{Advances in Neural Information Processing Systems 30: Annual
  Conference on Neural Information Processing Systems 2017, December 4-9, 2017,
  Long Beach, CA, {USA}}, pages 4077--4087.

\bibitem[{Tjong Kim~Sang(2002)}]{tjong2003introduction}
Erik~F. Tjong Kim~Sang. 2002.
\newblock \href {https://aclanthology.org/W02-2024} {Introduction to the
  {C}o{NLL}-2002 shared task: Language-independent named entity recognition}.
\newblock In \emph{{COLING}-02: The 6th Conference on Natural Language Learning
  2002 ({C}o{NLL}-2002)}.

\bibitem[{Tong et~al.(2021)Tong, Wang, Xu, Cao, Liu, Hou, and
  Li}]{tong2021fewother}
Meihan Tong, Shuai Wang, Bin Xu, Yixin Cao, Minghui Liu, Lei Hou, and Juanzi
  Li. 2021.
\newblock \href {https://doi.org/10.18653/v1/2021.acl-long.487} {Learning from
  miscellaneous other-class words for few-shot named entity recognition}.
\newblock In \emph{Proceedings of the 59th Annual Meeting of the Association
  for Computational Linguistics and the 11th International Joint Conference on
  Natural Language Processing (Volume 1: Long Papers)}, pages 6236--6247,
  Online. Association for Computational Linguistics.

\bibitem[{Triantafillou et~al.(2020)Triantafillou, Zhu, Dumoulin, Lamblin,
  Evci, Xu, Goroshin, Gelada, Swersky, Manzagol, and
  Larochelle}]{trian2020metadata}
Eleni Triantafillou, Tyler Zhu, Vincent Dumoulin, Pascal Lamblin, Utku Evci,
  Kelvin Xu, Ross Goroshin, Carles Gelada, Kevin Swersky, Pierre{-}Antoine
  Manzagol, and Hugo Larochelle. 2020.
\newblock \href {https://openreview.net/forum?id=rkgAGAVKPr} {Meta-dataset: {A}
  dataset of datasets for learning to learn from few examples}.
\newblock In \emph{8th International Conference on Learning Representations,
  {ICLR} 2020, Addis Ababa, Ethiopia, April 26-30, 2020}. OpenReview.net.

\bibitem[{van~der Maaten and Hinton(2008)}]{van2008tsne}
Laurens van~der Maaten and Geoffrey Hinton. 2008.
\newblock \href {http://jmlr.org/papers/v9/vandermaaten08a.html} {Visualizing
  data using t-sne}.
\newblock \emph{Journal of Machine Learning Research}, 9(86):2579--2605.

\bibitem[{Vinyals et~al.(2016)Vinyals, Blundell, Lillicrap, Kavukcuoglu, and
  Wierstra}]{vinyal-2016-matchnet}
Oriol Vinyals, Charles Blundell, Tim Lillicrap, Koray Kavukcuoglu, and Daan
  Wierstra. 2016.
\newblock \href
  {https://proceedings.neurips.cc/paper/2016/hash/90e1357833654983612fb05e3ec9148c-Abstract.html}
  {Matching networks for one shot learning}.
\newblock In \emph{Advances in Neural Information Processing Systems 29: Annual
  Conference on Neural Information Processing Systems 2016, December 5-10,
  2016, Barcelona, Spain}, pages 3630--3638.

\bibitem[{Wang et~al.(2021{\natexlab{a}})Wang, Xu, Liu, Zhou, Cao, Chang, and
  Sui}]{wang2021enhanced}
Peiyi Wang, Runxin Xu, Tianyu Liu, Qingyu Zhou, Yunbo Cao, Baobao Chang, and
  Zhifang Sui. 2021{\natexlab{a}}.
\newblock \href {https://arxiv.org/abs/2109.13023} {An enhanced span-based
  decomposition method for few-shot sequence labeling}.
\newblock \emph{ArXiv preprint}, abs/2109.13023.

\bibitem[{Wang et~al.(2021{\natexlab{b}})Wang, Chu, Zhang, and
  Gao}]{wang2021learning}
Yaqing Wang, Haoda Chu, Chao Zhang, and Jing Gao. 2021{\natexlab{b}}.
\newblock Learning from language description: Low-shot named entity recognition
  via decomposed framework.
\newblock In \emph{Findings of the Association for Computational Linguistics:
  EMNLP 2021}.

\bibitem[{Wolf et~al.(2020)Wolf, Debut, Sanh, Chaumond, Delangue, Moi, Cistac,
  Rault, Louf, Funtowicz, Davison, Shleifer, von Platen, Ma, Jernite, Plu, Xu,
  Le~Scao, Gugger, Drame, Lhoest, and Rush}]{wolf2020transformers}
Thomas Wolf, Lysandre Debut, Victor Sanh, Julien Chaumond, Clement Delangue,
  Anthony Moi, Pierric Cistac, Tim Rault, Remi Louf, Morgan Funtowicz, Joe
  Davison, Sam Shleifer, Patrick von Platen, Clara Ma, Yacine Jernite, Julien
  Plu, Canwen Xu, Teven Le~Scao, Sylvain Gugger, Mariama Drame, Quentin Lhoest,
  and Alexander Rush. 2020.
\newblock \href {https://doi.org/10.18653/v1/2020.emnlp-demos.6} {Transformers:
  State-of-the-art natural language processing}.
\newblock In \emph{Proceedings of the 2020 Conference on Empirical Methods in
  Natural Language Processing: System Demonstrations}, pages 38--45, Online.
  Association for Computational Linguistics.

\bibitem[{Wu et~al.(2020)Wu, Lin, Wang, Chen, Karlsson, Huang, and
  Lin}]{wu2020enhanced}
Qianhui Wu, Zijia Lin, Guoxin Wang, Hui Chen, B{\"{o}}rje~F. Karlsson, Biqing
  Huang, and Chin{-}Yew Lin. 2020.
\newblock \href {https://aaai.org/ojs/index.php/AAAI/article/view/6466}
  {Enhanced meta-learning for cross-lingual named entity recognition with
  minimal resources}.
\newblock In \emph{The Thirty-Fourth {AAAI} Conference on Artificial
  Intelligence, {AAAI} 2020, The Thirty-Second Innovative Applications of
  Artificial Intelligence Conference, {IAAI} 2020, The Tenth {AAAI} Symposium
  on Educational Advances in Artificial Intelligence, {EAAI} 2020, New York,
  NY, USA, February 7-12, 2020}, pages 9274--9281. {AAAI} Press.

\bibitem[{Yan et~al.(2021)Yan, Gui, Dai, Guo, Zhang, and
  Qiu}]{yan-etal-2021-ner-generative}
Hang Yan, Tao Gui, Junqi Dai, Qipeng Guo, Zheng Zhang, and Xipeng Qiu. 2021.
\newblock \href {https://doi.org/10.18653/v1/2021.acl-long.451} {A unified
  generative framework for various {NER} subtasks}.
\newblock In \emph{Proceedings of the 59th Annual Meeting of the Association
  for Computational Linguistics and the 11th International Joint Conference on
  Natural Language Processing (Volume 1: Long Papers)}, pages 5808--5822,
  Online. Association for Computational Linguistics.

\bibitem[{Yang and Katiyar(2020)}]{yang2020simple}
Yi~Yang and Arzoo Katiyar. 2020.
\newblock \href {https://doi.org/10.18653/v1/2020.emnlp-main.516} {Simple and
  effective few-shot named entity recognition with structured nearest neighbor
  learning}.
\newblock In \emph{Proceedings of the 2020 Conference on Empirical Methods in
  Natural Language Processing (EMNLP)}, pages 6365--6375, Online. Association
  for Computational Linguistics.

\bibitem[{Yoon et~al.(2019)Yoon, Seo, and Moon}]{yoon-2019-tapnet}
Sung~Whan Yoon, Jun Seo, and Jaekyun Moon. 2019.
\newblock \href {http://proceedings.mlr.press/v97/yoon19a.html} {Tapnet: Neural
  network augmented with task-adaptive projection for few-shot learning}.
\newblock In \emph{Proceedings of the 36th International Conference on Machine
  Learning, {ICML} 2019, 9-15 June 2019, Long Beach, California, {USA}},
  volume~97 of \emph{Proceedings of Machine Learning Research}, pages
  7115--7123. {PMLR}.

\bibitem[{Yu et~al.(2021)Yu, He, Zhang, Du, Pasupat, and Li}]{yu2021fewShot}
Dian Yu, Luheng He, Yuan Zhang, Xinya Du, Panupong Pasupat, and Qi~Li. 2021.
\newblock \href {https://doi.org/10.18653/v1/2021.naacl-main.59} {Few-shot
  intent classification and slot filling with retrieved examples}.
\newblock In \emph{Proceedings of the 2021 Conference of the North American
  Chapter of the Association for Computational Linguistics: Human Language
  Technologies}, pages 734--749, Online. Association for Computational
  Linguistics.

\bibitem[{Zeldes(2017)}]{zeldes2017gum}
Amir Zeldes. 2017.
\newblock The gum corpus: Creating multilayer resources in the classroom.
\newblock \emph{Language Resources and Evaluation}, 51(3):581--612.

\end{thebibliography}

\newpage
\appendix


\counterwithin{figure}{section}
\counterwithin{table}{section}
\section{Appendix}



\subsection{Meta learning}
\label{sec:meta}
The goal of meta-learning is to learn to fast adapt to a new few-shot task that is never-seen-before. To train a meta-learning model, a large number of episodes $\mathcal{T}_{train}$ (few-shot tasks) are constructed from training data $D_{train}$, which usually follows the $N$-way $K$-shot task formulation and are used to train the meta-learning model. One episode contains a small training set $\mathcal{S}_{train}$, called support set, and a test set $\mathcal{Q}_{train}$, called query set. The meta-learner generates a task-specific model for a new task $\mathcal{T}_i$ via updating on support set $\mathcal{S}_{train}$, then the task-specific model is tested on $\mathcal{Q}_{train}$ to get a test error. The meta-learner then learns to learn new tasks by considering how to reduce the test error on $\mathcal{Q}_{train}$ by updating on $\mathcal{S}_{train}$.  To evaluate the task learning ability of a meta-learner, a bunch of episodes $\mathcal{T}_{test}$ are constructed from the normal test data $D_{test}$, and the expectation of performance on $\mathcal{Q}_{test}$ from all test episodes is severed as evaluation protocol. To distinguish the training phase of meta-learner on episodes $\mathcal{T}_{train}$ and training of a task-specific model on support set $\mathcal{S}$, the former is called meta-training and the latter is called training. Similarly, the testing of a meta-learner on $\mathcal{T}_{test}$ is called meta-testing, and the evaluating of a task-specific model on query set $\mathcal{Q}$ is called testing.



\subsection{Datasets}
\label{sec:dataset}
Table \ref{tab:dataset-statistic} shows the dataset statistics of original data for constructing few-shot episodes. 

\begin{table}[htbp]
    \centering
    \small
    \begin{tabular}{cccc}
    \toprule
       \textbf{Dataset}  & \textbf{Domain} & \textbf{\# Sentences} & \textbf{\# Classes} \\
     \midrule
       Few-NERD  & Wikipedia & 188.2k & 66 \\
       CoNLL03 & News & 20.7k & 4 \\
       GUM & Wiki & 3.5k & 11 \\
       WNUT & Social & 5.6k & 6 \\
       OntoNotes & Mixed & 159.6k & 18 \\
     \bottomrule
    \end{tabular}
    \caption{Evaluation dataset statistics}
    \label{tab:dataset-statistic}
\end{table}

For Few-NERD, we use episodes released by \citet{ding2021nerd}\footnote{\url{https://ningding97.github.io/fewnerd/}} which contain 20,000 episodes for training, 1,000 episodes for validation, and 5,000 episodes for testing. Each episode is an N-way K$\sim$2K-shot few-shot task. As for Cross-Dataset, 
two datasets are used for constructing training episodes, one dataset is used for validation, and episodes from the remained dataset are used for evaluation. We use public episodes\footnote{\url{https://github.com/AtmaHou/FewShotTagging}} constructed by \citet{hou2020few}. For 5shot, 200 episodes are used for training, 100 episodes for validation, and 100 for testing. For the 1shot experiment, 400/100/200 episodes are used for training/validation/testing, except for experiments on OntoNotes(Mixed), where 400/200/100 episodes are constructed for train/dev/test.

\subsection{Additional Implementation Details}
\label{sec:details}
\paragraph{Parameter Setting}

We use \texttt{BERT-base-unca sed} from Huggingface Library \citep{wolf2020transformers} as our base encoder following \citet{ding2021nerd}. We use AdamW \citep{loshchilov2018decoupled} as our optimizer with a learning rate of 3e-5 and 1\% linear warmup steps at both the meta-training and finetuning in meta-testing time for all experiments. The batch size is set to 32, the max sequence length is set to 128 and we keep dropout rate as 0.1. At meta-training phase, the inner update step is set to 2 for all experiments. When finetuning the span detector at meta-testing phase, the finetune step is set to 3 for all inter settings on Few-NERD dataset and 30 for other experiments. For entity typing, the finetune step at meta-testing phase is set to 3 for all experiments on Few-NERD dataset, 20 for all Cross-Dataset experiments. To further boost the performance, we only keep entities that have a similarity score with its nearest prototype greater than a threshold of 2.5. We set max-loss coefficient $\lambda$ as 2 at meta-training query set evaluation phase, 5 at other phases. We validate our model on dev set every 100 steps and select the checkpoint with best f1 score performance on dev set within the max train steps 1,000. We use grid search for hyperparameter setting, the search space is shown in Table \ref{tab:hyper-parameters}. The total model has 196M parameters and trains in $\approx$60min on a Tesla V100 GPU.

\begin{table}[htb]
\centering
\small
\begin{tabular}{lc}
\toprule
Learning rate & \{1e-5, 3e-5, 1e-4\} \\
Meta-test fine-tune steps & \{3, 5, 10, 20, 30\} \\
Max-loss coefficient $\lambda$ & \{0, 1, 2, 5, 10\} \\
Type similarity threshold & \{1, 2.5, 5\} \\
Mini-batch size & \{16, 32\} \\ 
\bottomrule
\end{tabular}
\caption{Hyper-parameters search space used in our experiments.}
\label{tab:hyper-parameters}
\end{table}

\subsection{Baselines}
\label{ref:baselines}
We consider the following metric-learning based baselines:

\textbf{SimBERT} \citep{hou2020few} applies BERT without any finetuning as the embedding function, then assign each token's label by retrieving the most similar token in the support set .

\textbf{ProtoBERT} \citep{fritzler2019few} uses a token-level prototypical network \citep{snell2017proto} which represents each class by averaging token representation with the same label, then the label of each token in the query set is decided by its nearest class prototype.

\textbf{MatchingBERT} \citep{vinyal-2016-matchnet} is similar to ProtoBERT except that it calculates the similarity between query instances and support instances instead of class prototypes.

\textbf{L-TapNet+CDT} \citep{hou2020few} enhances TapNet \citep{yoon-2019-tapnet} with pair-wise embedding, label semantic, and CDT transition mechanism. 

\textbf{NNShot} \citep{yang2020simple} pretrains BERT for token embedding by conventional classification for training, a token-level nearest neighbor method is used at testing.

\textbf{StructShot} \citep{yang2020simple} improves NNshot by using an abstract transition probability for Viterbi decoding at testing.

\textbf{ESD} \citep{wang2021enhanced} is a span-level metric learning based method. It enhances prototypical network by using inter- and cross-span attention for better span representation and designs multiple prototypes for O label. 

Besides, we also compare with the finetune-based methods:

\textbf{TransferBERT} \citep{hou2020few} trains a token-level BERT classifier, then finetune task-specific linear classifier on support set at test time.

\textbf{CONTAINER} \citep{das2021container} uses token-level contrastive learning for training BERT as token embedding function, then finetune the BERT on support set and apply a nearest neighbor method at inference time.

\subsection{Results of Span Detection}
Table \ref{tab:performance_comparison_fewnerd_span} and Table \ref{tab:performance_comparison_crossdataset_span} show the performance of our span detection module on Few-NERD and Cross-Dataset.

\label{sec:results_span}
\begin{table}[t]
    \centering
    \setlength{\tabcolsep}{1mm}
    \resizebox{\columnwidth}{!}{
    \begin{tabular}{ccccc}
    \toprule
        \multirow{2}{*}{\textbf{Models}} 
        & \multicolumn{2}{c}{\textbf{1$\sim$2-shot}} & \multicolumn{2}{c}{\textbf{5$\sim$10-shot}}\\
        \cmidrule(lr){2-3}\cmidrule(lr){4-5}
         & 5 way & 10 way & 5 way & 10 way \\
         \midrule
         \textbf{Intra} & 73.69\small{\textpm0.14} & 74.32\small{\textpm1.84} & 77.76\small{\textpm0.24} & 78.66\small{\textpm0.15}\\
         \textbf{Inter} & 76.71\small{\textpm0.30} & 76.63\small{\textpm0.24} & 75.97\small{\textpm0.14} & 76.62\small{\textpm0.11} \\
        \bottomrule
    \end{tabular}
    }
    \caption{F1 scores of our entity span detection module on \textbf{Few-NERD} for both inter and intra settings.}
    \label{tab:performance_comparison_fewnerd_span}
\end{table}

\begin{table}[t]
    \centering
    \setlength{\tabcolsep}{1mm}
    \resizebox{\columnwidth}{!}{
    \begin{tabular}{ccccc}
        \toprule
         \textbf{Models} & \textbf{News} & \textbf{Wiki} & \textbf{Social} & \textbf{Mixed}\\
         \midrule
         \textbf{1-shot} & 65.06\small{\textpm0.91} & 35.63\small{\textpm2.17} & 38.89\small{\textpm0.55} & 46.52\small{\textpm1.24} \\
          \textbf{5-shot} & 74.20\small{\textpm0.33} & 46.26\small{\textpm1.28} & 43.16\small{\textpm1.23} & 54.70\small{\textpm0.88}\\
        \bottomrule
    \end{tabular}
    }
    \caption{F1 scores of our entity span detection module on \textbf{Cross-Dataset}.}
    \label{tab:performance_comparison_crossdataset_span}
\end{table}
\end{document}